\documentclass[11pt]{article}
\usepackage{paper}
\usepackage{times}
\usepackage{latexsym}
\usepackage{graphicx,topcapt}
\usepackage{numprint}
\usepackage{amsmath}
\usepackage{color}
\usepackage{url}

\aclfinalcopy 

\title{Neural Morphological Tagging from Characters\\for Morphologically Rich Languages}

\author{Georg Heigold \and G\"unter Neumann \and Josef van Genabith\\
	    DFKI \& Saarland University\\
	    Stuhlsatzenhausweg 3\\
	    66123 Saarbr\"ucken, Germany\\
	    {\tt <firstname>.<lastname>@dfki.de}
  }

\date{}

\begin{document}

\maketitle

\begin{abstract}
This paper investigates neural character-based morphological tagging for languages with complex morphology and large tag sets. 
We systematically explore a variety of neural architectures (DNN, CNN, CNNHighway, LSTM, BLSTM)
to obtain character-based word vectors combined with bidirectional LSTMs to model across-word context in an end-to-end setting. 
We explore supplementary use of word-based vectors trained on large amounts of unlabeled data. 
Our experiments for morphological tagging suggest that
for "simple" model configurations, the choice of the network architecture (CNN vs. CNNHighway vs. LSTM vs. BLSTM)
or the augmentation with pre-trained word embeddings can be important and clearly impact the accuracy.
Increasing the model capacity by adding depth, for example, and carefully optimizing the neural networks can lead
to substantial improvements, and the differences in accuracy (but not training time) become much smaller or even negligible. 
Overall, our best morphological taggers for German and Czech outperform the best results reported in the literature by a large margin.
\end{abstract}

\section{Introduction}
\begin{figure*}[htbp] 
   \centering
   \includegraphics[width=1.0\textwidth]{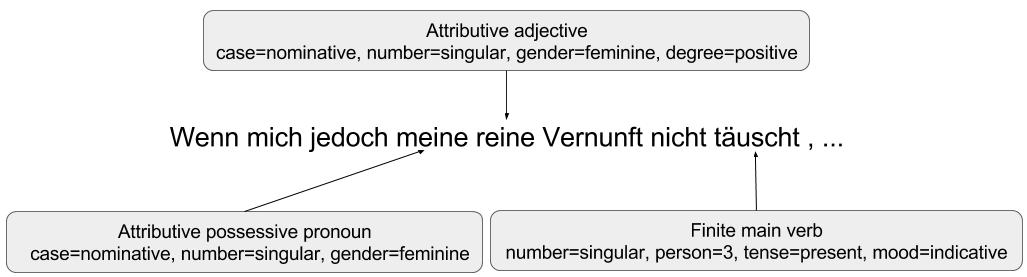} 
   \caption{Example for morphological tagging: The first line in the gray boxes with the part of speech maps to the POS tag,
   the second line with the morphological information maps to the MORPH tag, the combination of POS tag and MORPH tag gives a single tag
   POSMORPH used for full morphological tagging.}
   \label{fig:tagging}
\end{figure*}

Morphological part-of-speech tagging is the process of marking up a word in a text 
with its morphological information and part of speech (POS), see Fig.~\ref{fig:tagging}.
In morphologically rich languages (e.g., Turkish and Finnish), 
individual words encode substantial amounts of grammatical information 
(such as number, person, case, gender, tense, aspect, etc.) in the word form, 
whereas morphologically poor languages (e.g., English) rely more on word order and context to express this information. 
Most languages (such as German and Czech) lie between these two extremes, and
some (e.g. German) exhibit syncretism, that is one-to-many mappings between form and function.
For example (Fig.~\ref{fig:tagging}), the isolated word form "meine" can be any combination of
"case=[nominative$|$accusative], number=[singular$|$plural], gender=feminine" (among others) 
of the possessive pronoun "my" or a verb ("to mean"). 
This suggests that both within and across word modeling is needed in general.

Morphologically rich languages exhibit large vocabulary sizes and
relatively high out-of-vocabulary (OOV) rates on the word level.
Table~\ref{tab:oov} illustrates this for German (TIGER, de.wikidump) and Czech (PDT).
\begin{table}[htbp]
   \centering
   \topcaption{Coverage of test data for different amounts of training data and numbers of training occurrences} 
   \begin{tabular}{@{} lrrrr @{}} 
      \hline
      & \#Tokens & \multicolumn{3}{c}{\#Occurrences}\\
       &  & 0 & 1-4 & $\geq$5\\
      \hline
      TIGER, train & 720k & 9.5\% & 7.7\% & 82.8\%\\
      de.wikidump & 610M & 3.6\% & 1.8\% & 94.5\%\\
      \hline
     PDT, train & 653k & 8.7\% & 10.4\% & 80.9\%\\
      \hline
   \end{tabular}
   \label{tab:oov}
\end{table}
Word-level representations generalize poorly to rarely seen or unseen words and thus,
can significantly impair the performance for high OOV rates.
To improve generalization, sub-word representations have been proposed.
Compared to morphemes as the sub-word unit~\cite{Luong:2013}, characters have the advantage
of being directly available from the text and do not require additional resources and complex pre-processing steps.
Character-based approaches may also be useful for informal language (e.g., Tweets) or low-resource languages.

This paper investigates character-based morphological tagging. 
More specifically, we
(i) provide a systematic evaluation of different neural architectures (DNN, CNN, CNNHighway, LSTM, BLSTM) to obtain 
   character-based word vectors (Table~~\ref{tab:inputnns}),
(ii) explore the supplementary use of word (rather than character-based) embeddings 
     pre-trained on large amounts of unlabeled data (Table~\ref{tab:word2vec}), and
(iii) show that carefully optimized character-based systems can outperform existing systems by a large margin for 
German (Table~\ref{tab:german}) and Czech (Table~\ref{tab:czech}).

The focus of the paper is to gain a better understanding of the relative importance of different basic neural architectures and building blocks.
Data from morphologically rich languages are well suited to amplify these differences as a large amount of 
relevant information is encoded at the word level along with relatively high OOV rates. This helps us to better distinguish between systematic trends and noise when comparing different neural architectures.
Similarly, we focus on morphological tagging, which typically has an order of magnitude more tags including also	
(where tags consist of sequences of a simple POS tag followed by many morphological feature=value pairs, see Fig.~\ref{fig:tagging})
 than simple POS tagging and correspondingly higher error rates.

The remainder of the paper is organized as follows.
Section~\ref{sec:soa} gives a survey on related work.
Section~\ref{sec:char-tagging} describes the neural network-based approach as explored in this paper.
The empirical evaluation is presented in Section~\ref{sec:experiments}.
Section~\ref{sec:summary} concludes the paper.

\section{Related Work}\label{sec:soa}
This work is in the spirit of the "natural language processing (almost) from scratch" approach~\cite{Collobert:2011b},
which was tested for word-level processing and various English natural language processing tasks.
Several character-based approaches have been proposed for tagging.
Existing work for POS tagging includes 
feature learning using CNNs in combination with a first-order Markov 
model for classification~\cite{Collobert:2011b,Santos:2014} and
recurrent neural network based approaches used in~\cite{Ling:2015a,Gillick:2015,Plank:2016}.
The work by~\cite{Labeau:2015} uses a CNN/Markov model hybrid for morphological tagging of German.
Comprehensive work on morphological tagging based on conditional random fields along with
state-of-the-art results can be found in~\cite{Mueller:2013,Mueller:2015}.
Our work is inspired by previous work~\cite{Collobert:2011b,Santos:2014,Labeau:2015,Ling:2015a} 
but uses a deeper hierarchy of layers in combination with a simple prediction model and
provides comprehensive comparative results for alternative neural architectures for morphological tagging.

Several extensions of the neural approach used in this paper have been proposed,
including multilingual training~\cite{Gillick:2015}, auxiliary tasks~\cite{Plank:2016}, 
and more structured prediction models~\cite{Collobert:2011b,Santos:2014,Labeau:2015,Ma&Hovy:2016}.
It is conceivable that these refinements would lead to further improvements.
In this paper, we focus on optimizing and comparing a number of architectures for obtaining character vectors
and try to keep the rest as simple as possible.

Character-based approaches have also been applied to other tasks in natural language processing, 
such as named entity recognition~\cite{Gillick:2015},
parsing~\cite{Ballesteros:2015} (BLSTM),
language modeling~\cite{Ling:2015a} (BLSTM) and \cite{Kim:2016} (CNNs) or
neural machine translation~\cite{Costa:2016}.

%
%
%

\section{From Characters to Tags}\label{sec:char-tagging}
We assume an input sentence $w_1,\dots,w_N$ with (possibly complex morphological)  
output tags $t_1,\dots,t_N$. We use a zeroth-order Markov model
\begin{eqnarray}\label{eq:condp}
  & p(t_1,\dots,t_N|w_1,\dots,w_N) \hspace{15ex} & \nonumber\\
  & \hspace{15ex} = \prod\limits_{n=1}^{N}p(t_n|w_1,\dots,w_N) &
  \end{eqnarray}
whose factors are modeled by a neural network.
When mapping characters to tags, we use the character representation of the word, $w=c_1,\dots,c_M$.
This assumes that the segmentation of the sentence into words is known, which 
is straightforward for the languages under consideration.

\begin{figure}[htbp] 
   \centering
   \includegraphics[width=0.5\textwidth]{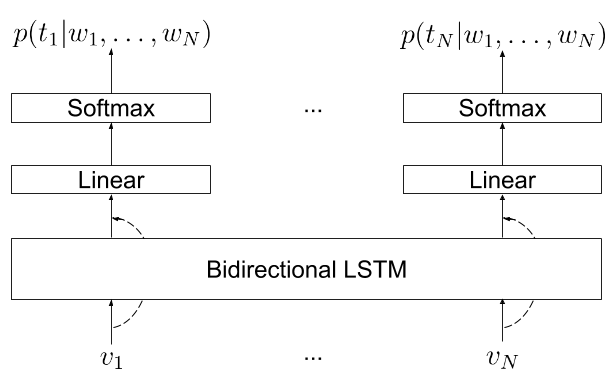} 
   \caption{Architecture mapping word vectors $v_1,\dots,v_N$ to tags $t_1,\dots,t_N$, 
   dashed arrows indicate optional skip connections.}
   \label{fig:inputnn-blstm}
\end{figure}

Each input word maps to one complex output tag. Hence, we can model the
position-wise probabilities $p(t|w_1,\dots,w_N)$ with recurrent neural networks, 
such as long short-term memory recurrent neural networks (LSTMs)~\cite{Graves:2012}.
Fig.~\ref{fig:inputnn-blstm} shows such a network architecture where the inputs are 
the word vectors $v_1,\dots,v_N$.
On top of the BLSTM, we use position-wise softmax classifiers. 

Fig.~\ref{fig:inputnn-blstm} shows the "upper" part of our network. 
This part is used in all experiments reported below. 
We now turn to the "lower" parts of our networks, 
where we experiment with different architectures to obtain character vectors that make up the $v_i$s. 
In fact, in our work, we use both word-based and character-based word vectors.
Word-based word vectors are attractive because they can be efficiently pre-trained
on supplementary, large amounts of unsupervised data~\cite{Mikolov:2013}.
As shown by~\cite{Soricut:2015}, these word vectors also encode morphological information
and may provide additional information to the character-based word vectors
directly learned from the comparably small amounts of supervised data.
We use word-based word vectors in two modes: they are pre-trained and kept fixed during
training or jointly optimized with the rest of the network.
Word-based vectors are efficient as they can be implemented by a lookup table (LUT)
(Fig.~\ref{fig:inputnns}, (a)) but are bad at generalization because they do not exploit
information encoded at the sub-word level.

The character-based word vectors are the output vectors of a sub-network
that maps variable-length character strings to fixed-length vectors.
In this paper, we compare the following mostly well established network architectures, see also Fig.~\ref{fig:inputnns}:
\begin{figure*}[htbp] 
   \centering
   \includegraphics[width=1.0\textwidth]{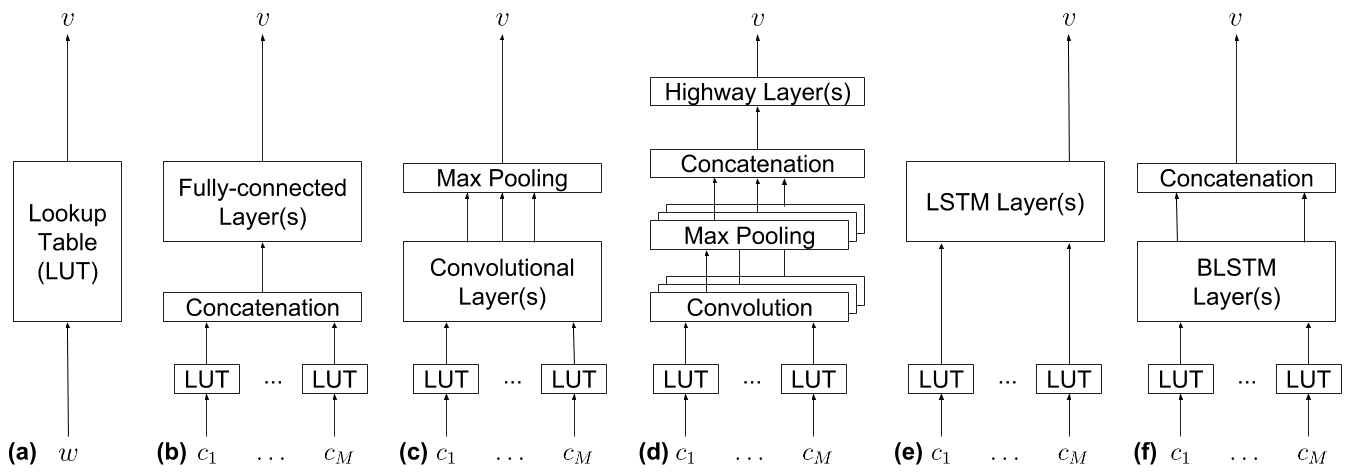} 
   \caption{Existing architectures mapping words and characters into word vectors $v$, either directly from word $w$ (a) or from the character string $c_1,\dots,c_M$ representing the word (b) - (f):
   (a) sparse linear transformation implemented by a lookup table (LUT), 
   (b) fully-connected deep neural network (DNN), 
   (c) convolutional neural network (CNN),
   (d) CNNs with different filter widths followed by fully-connected layers with highway connections (CNNHighway),
   (e) deep LSTM using the last output for the character-based word vector, and 
   (f) deep BLSTM using the last outputs from the forward and backward LSTMs.}
   \label{fig:inputnns}
\end{figure*}
\begin{itemize}
\item Fully-connected deep neural networks (DNNs): 
  DNNs expect fixed-length input vectors.
  To satisfy this constraint, we assume a maximum number of characters per word.
  Fixed-length character strings from words are obtained by padding with a special character.
  The fixed-length sequence of character vectors can then be converted into a fixed-length vector 
  by concatenation, which is fed to the DNN.
  DNNs are generic, unstructured networks which tend to be inefficient to learn in general.
\item Convolutional neural networks (CNNs)~\cite{Collobert:2011b,Santos:2014,Labeau:2015}: 
  Compared to DNNs, CNNs use weight tying and local connections.
  This makes CNNs more efficient to learn in many settings.
  CNNs can deal with variable-length input across different batches and
  produce a variable number of output vectors, which are merged into a single fixed-length vector by max pooling.
  The context length is controlled by the pre-defined filter width.
  For a filter width of $m$, a convolutional layer computes ("hierarchical" in case of multiple layers) character m-gram vectors.
  CNNs scale well to long sequences and are efficient due to highly optimized libraries.
\item CNNHighway~\cite{Kim:2016,Costa:2016}: 
  This CNN variant is similar to vanilla CNNs but maintains a set of one-layer CNNs with different filter widths.
  This alleviates problems with having a single filter width and selecting an appropriate filter width.
  The outputs of the different CNNs are concatenated and max pooled, followed by a fully-connected
  deep neural network with highway connections for additional mixing.
  CNNHighway includes many layers but is basically a shallow architecture, which tends to make learning easier.
\item LSTMs~\cite{Ling:2015a}:
  LSTMs are sequential models and thus a natural choice for character strings.
  Vanilla LSTMs map each input to one output.
  To obtain a fixed-length vector, only the last output vector, ideally encoding the whole sequence, 
  is used as the word vector; all other outputs are suppressed.
  Unlike CNNs, recurrent neural networks can learn context of variable length and do not use a pre-defined context length.
  In general, multiple layers are required to perform the complex transformations.
  A disadvantage of deep LSTMs is that they can be difficult to train.
\item Bidirectional LSTMs (BLSTMs)~\cite{Ling:2015a,Ballesteros:2015,Plank:2016}:
   BLSTMs are similar to LSTMs but encode the input from left to right and from right to left.
   The word vector is the concatenation of the output vector of the (topmost) forward LSTM
   at position $M$ and the output vector of the (topmost) backward LSTM at position $1$.
   For a "perfect" sequence model, it might not be obvious why the word needs to be encoded
   in both directions.
\end{itemize}
Where applicable, word-level and character-level word vectors are combined by concatenation.

The weights of the network, $\theta$, are jointly estimated using the conditional log-likelihood
\begin{equation}\label{eq:ce}
  F(\theta) = -\sum_{n=1}^{N}\log p_{\theta}(t_n|w_1,\dots,w_N).
\end{equation}
Learning in recurrent or very deep neural networks is non-trivial and 
skip/shortcut connections have been proposed to improve the learning of such networks~\cite{Pascanu:2014,He:2016}.
Here, we use such connections (dashed arrows in Fig.~\ref{fig:inputnn-blstm}) 
in some of the experiments to alleviate potential learning issues.

At test time, the predicted tag sequence is the tag sequence 
that maximizes the conditional probability $p(t_1,\dots,t_N|w_1,\dots,w_N)$.
For the factorization in Eq.~(\ref{eq:condp}), the search can be done position-wise.
This significantly reduces the computational and implementation complexity compared to first-order Markov 
models as used in~\cite{Collobert:2011b,Santos:2014,Labeau:2015}.

\section{Experimental Results}\label{sec:experiments}
We first test variants of the architecture for German (Section~\ref{sec:german}) 
and then verify our empirical findings for Czech (Section~\ref{sec:czech}).

\subsection{Data}\label{sec:data}
We conduct the experiments on the German TIGER 
corpus\footnote{\url{http://www.ims.uni-stuttgart.de/forschung/ressourcen/korpora/tiger.html}} 
and the Czech PDT corpus\footnote{\url{https://ufal.mff.cuni.cz/pdt3.0}}. 
For the time being, 
we have decided against using the recent Universal Dependencies~\footnote{\url{http://universaldependencies.org/}}
because of the lack of comparative results for morphological tagging in the literature.
Table~\ref{tab:oov} (at the beginning of the paper) 
presents OOV rates and Table~\ref{tab:data} some corpus statistics.
\begin{table}[htbp]
   \centering
   \topcaption{Corpus statistics} 
   \begin{tabular}{@{} llrr @{}} 
      \hline
      Language & Data set & \#Sentences & \#Words\\
      \hline
      German & de.wikidump & 36M & 610M\\
      & TIGER, train & 40,474 & 719,530\\
      & \phantom{TIGER, }dev & 5,000 & 76,704\\
      & \phantom{TIGER, }test & 5,000 & 92,004\\
      \hline
      Czech & cs.wikidump & 5M & 83M\\
      & PDT, train & 38,727 & 652,544\\
      & \phantom{PDT, }test & 4,213 & 70,348\\
      \hline
   \end{tabular}
   \label{tab:data}
\end{table}
Part of the experiments is supervised learning on small labeled data sets (TIGER, PDT)
and part is also including large unlabeled data (de.wikidump, cs.wikidump)\footnote{\url{http://cistern.cis.lmu.
de/marmot/}}.
The tag set sizes observed in the labeled training data depend on the language and the type of tags: 
54 (POS, German), 255 (MORPH, German), 681 (POSMORPH, German), and 1,811 (POSMORPH, Czech),
where POS stands for the part-of-speech tags, MORPH for the morphological tags (feature=value pairs), 
and POSMORPH for the combined tag sets POS and MORPH.
All words are lowercased.
As a result of doing this, we ignore a useful hint for nouns in German 
(which makes a difference in error rate for the simple but not for the best models)
but makes the conclusions less dependent on this German-specific feature.

\subsection{Setup}

We empirically tuned the hyperparameters on the TIGER development data and used the same setups also for Czech.
The best setups for the character-based word vector neural networks are as follows:
\begin{itemize}
\item DNN: character vector size = 128, one fully-connected layer with 256 hidden nodes 
\item CNN: character vector size = 128, two convolutional layers with 256 filters and a filter width of five each 
\item CNNHighway: the large setup from~\cite{Kim:2016}, i.e., 
  character vector size = 15, filter widths ranging from one to seven,
  number of filters as a function of the filter width $\min\{200,50\cdot\text{filter width}\}$, 
  two highway layers
\item LSTM: character vector size = 128, two layers with 1024 and 256 nodes
\item BLSTM: character vector size = 128, two layers with 256 nodes each
\end{itemize}
The BLSTM modeling the context of words in a sentence (Fig.~\ref{fig:inputnn-blstm}) 
consists of two hidden layers, each with 256 hidden nodes.

The training criterion in Eq.~(\ref{eq:ce}) is optimized using standard backpropagation 
and \mbox{RMSProp}~\cite{Tieleman:2012} with a learning rate decay of two every tenth epoch.
The batch size is 16. 
We use dropout on all parts of the networks except on the lookup tables to reduce overfitting. 
In particular and in contrast to~\cite{Zaremba:2015},
we also use dropout on the recurrent parts of the network because it gives significantly better results.
Training is stopped when the error rate on the development set has converged, which typically is after about 50 epochs.
We observe hardly any overfitting with the described configurations.

We used Torch~\cite{Collobert:2011a} to configure the computation graphs implementing the network architectures.

\subsection{German}\label{sec:german}
We first establish a baseline for German and compare it with the state of the art.
Our baseline model (CNN-BLSTM) consists of the BLSTM in Fig.~\ref{fig:inputnn-blstm} 
and the CNN in Fig.~\ref{fig:inputnns} (c) with a single convolutional layer, which
is a simplified version of the best model in~\cite{Labeau:2015}.
The results are summarized in Table~\ref{tab:german}.
We show results for different tag sets (see Section~\ref{sec:data}) to facilitate the comparison with
state-of-the-art results. 

Our CNN-BLSTM baseline achieves comparable or better results for all tag sets.
In particular, our CNN-BLSTM clearly (under consistent conditions) outperforms the related models in~\cite{Labeau:2015} 
and~\cite{Ling:2015a}\footnote{We downloaded the software from \url{https://github.com/wlin12/JNN} to
produce consistent results as the results in~\cite{Ling:2015a} are for the last 100 training sentences only and not for the standard test set. We use the default settings given in the paper for all experiments.}.
As expected, word-level word vectors on their own (Fig.~\ref{fig:inputnns} (a)) perform significantly worse
than character-level word vectors, with error rates of 5.76\% vs. 2.43\% for POS tagging.
Combining character-level and word-level word vectors computed on the TIGER training data only did not help.

Word embeddings~\cite{Mikolov:2013} or word clusters~\cite{Mueller:2015} allow us
to exploit large amounts of unlabeled data.
Using pre-trained word embeddings alone is better than the state of the art (9.27\% vs. 10.97\% for POSMORPH)
but does not improve on our results (9.27\% vs. 8.72\% for POSMORPH).
When combining them with the character-level word vectors, however,
large additional gains are observed: 2.43\% vs. 1.75\% for POS and 8.72\% vs. 6.67\% for POSMORPH
(Table~\ref{tab:german}).

Next, we compare different architectures to compute character-based word vectors for POSMORPH,
see Table~\ref{tab:inputnns}.
Note that here and in contrast to Table~\ref{tab:german}, we use multiple hidden layers in general,
which gives some additional gains.
We also tested whether skip connections as shown in Fig.~\ref{fig:inputnn-blstm} helps learning.
A small gain is observed only for LSTM, in all other cases it does not make a difference.
Given sufficient capacity (e.g., the number of hidden layers),
the different architectures achieve comparable error rates, except for the DNN which performs worse.
CNNHighway may perform slightly better than CNN.
CNN and CNNHighway are more memory efficient than LSTM and BLSTM
but considerably slower in our Torch-based configuration, for example,
0.5 sec/batch (BLSTM) vs. 2 sec/batch (CNNHighway). 
The only optimization we do here is to compute the word vectors of a batch in parallel.

\begin{figure}[htbp] 
   \centering
   \includegraphics[width=0.5\textwidth]{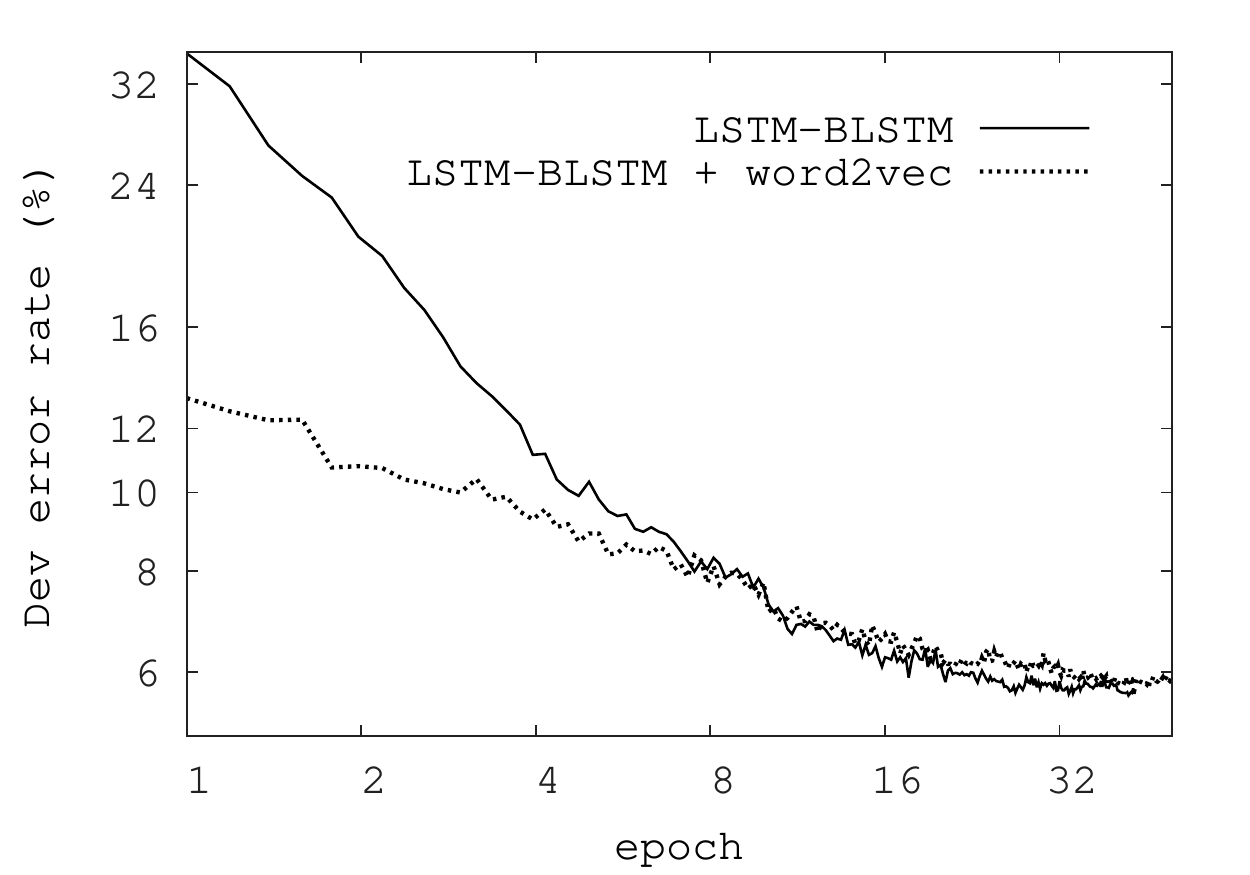} 
   \caption{Convergence curve for LSTM-BLSTM, without and with word2vec.}
   \label{fig:convergence}
\end{figure}

Finally, we investigate the effect of augmenting the character-based word vectors with
pre-trained word embeddings ("word2vec").
The gains for simpler models are promising: 8.72\% vs. 6.67\% for the one-layer CNN.
For more complex models, however, the observed gains are much smaller (6.77\% vs. 6.15\% for the best LSTM, for example).
Overall, the error rates without word2vec vary between 7\% and 9\% while the error rates with word2vec
are all around 7\%.
In particular, we cannot significantly improve over the best result in Table~\ref{tab:german} (6.67\% vs. 6.40\%).
In this example, the convolutional networks seem to better combine with word2vec than the recurrent neural network.
The convergence curve for LSTM-BLSTM augmented with word2vec on a subset of the development set is shown
in Fig.~\ref{fig:convergence}.
The initial convergence is faster with word2vec (ignoring the time to generate word2vec) but
the two curves eventually converge to the same error rate.

\subsection{Czech}\label{sec:czech}
We confirm our empirical findings for German on another morphologically rich language (Czech).
The results are summarized in Table~\ref{tab:czech} for the models that performed best on German.
Similar to German, CNNHighway-BLSTM and LSTM-BLSTM perform similarly (0.5\% absolute difference in error rate) 
and clearly better than the baselines (25\% or more relative error rate reduction).
Augmenting the character-based word vectors with pre-trained embeddings gives
some additional small gain. Again, the gain for CNNHighway-BLSTM is larger than for LSTM-BLSTM.

\section{Summary}\label{sec:summary}
This paper summarizes our empirical evaluation of the character-based 
neural network approach to morphological tagging.
Our empirical findings for German and Czech are as follows.
As long as carefully tuned neural networks of sufficient capacity 
(e.g., number of hidden layers) are used, 
the effect of the specific network architecture (e.g., convolutional vs. recurrent) is small
for the task under consideration.
However, the choice of architecture can greatly affect the training time
(in our implementation, the convolutional networks are 2-4 times slower than the recurrent networks).
Augmenting the character-based word vectors with word embeddings pre-trained on large amounts of unsupervised data,
gives large gains for the small configurations but only small gains on top of the best configurations.
Moreover, our best character-based morphological taggers outperform the state-of-the-art results 
for German and Czech by a relative gain of 30\% or more.
Future work will include the investigation of multilingual training, higher-order Markov models,
and low-resource languages.

\begin{table*}[p]
   \centering
   \topcaption{Test error rates (\%) for German, CNN for character-based word vectors has only one layer} 
   \npdecimalsign{.}
   \nprounddigits{2}
   \begin{tabular}{@{} llN{2}{2}N{2}{2}N{2}{2} @{}} 
      \hline
Train data & Setup & POS & MORPH &POSMORPH \\
      \hline
TIGER &      CRF~\cite{Mueller:2015} & 2.68 & 11.59 & \\      
  &    PCRF~\cite{Mueller:2013} & 2.56 & & 11.42 \\
&      biRNN, Non-Lex/Struct~\cite{Labeau:2015} & 3.59 & & 12.88 \\
&      biRNN, Both/Struct~\cite{Labeau:2015} & 2.86 & & 10.97 \\
  & BLSTM, lower-case~\cite{Ling:2015a}$^{5}$ & 3.07 & & 10.04\\
  & BLSTM, mixed case~\cite{Ling:2015a}$^{5}$ & 2.59 & & 9.24\\
      \cline{2-5}
&      CNN-BLSTM, word-based word vectors & 5.76 & &  \\
&      \phantom{CNN-BLSTM, }char-based word vectors & 2.43 & 7.98 & 8.72 \\
      \hline
+ de.wikidump &      CRF + MarLiN~\cite{Mueller:2015} & 2.27 & 10.82 & \\      
      \cline{2-5}
 &     CNN-BLSTM, word2vec & 2.61 & 8.55 & 9.27\\
&      \phantom{CNN-}char-based word vectors+word2vec & 1.75 & 6.16 & 6.67 \\
      \hline
   \end{tabular}
   \npnoround
   \label{tab:german}
\end{table*}

\begin{table*}[p]
   \centering
   \topcaption{Test error rates (\%) for German and different character-based word vectors} 
   \begin{tabular}{lrr} 
      \hline
 & \multicolumn{2}{c}{POSMORPH}\\
& BLSTM & + skip connection \\
      \hline
      CNN baseline~\cite{Labeau:2015} & 10.97 &\\
      BLSTM baseline~\cite{Ling:2015a}$^{5}$ & 10.04 &\\
      \hline
      DNN (Fig.~\ref{fig:inputnns} (b)) & 10.00 &\\
      CNN (Fig.~\ref{fig:inputnns} (c)) & 8.10 & 8.16\\
      CNNHighway (Fig.~\ref{fig:inputnns} (d)) & 7.37 & 7.20\\
      BLSTM (Fig.~\ref{fig:inputnns} (f)) & 7.50 & 7.53\\
      LSTM (Fig.~\ref{fig:inputnns} (e)) & 7.45 & 6.77\\
      \hline
   \end{tabular}
   \npnoround
   \label{tab:inputnns}
\end{table*}

\begin{table*}[p]
   \centering
   \topcaption{Test error rates (\%) for German and different character-based word vectors annotated with word2vec} 
   \begin{tabular}{lrrr} 
      \hline
      & \multicolumn{2}{c}{POSMORPH}\\
      & Char-based word vector & + word2vec \\
      \hline
      CNN, 1 layer (Fig.~\ref{fig:inputnns} (c)) & 8.72 & 6.67\\
      CNN, 2 layers (Fig.~\ref{fig:inputnns} (c)) & 8.10 & 6.66\\
      CNNHighway (Fig.~\ref{fig:inputnns} (d)) & 7.37 & 6.40\\
      LSTM (Fig.~\ref{fig:inputnns} (e)) & 6.77 & 6.15\\
      \hline
   \end{tabular}
   \npnoround
   \label{tab:word2vec}
\end{table*}

\begin{table*}[p]
   \centering
   \topcaption{Test error rates (\%) for Czech} 
   \npdecimalsign{.}
   \nprounddigits{2}
   \begin{tabular}{@{} llN{2}{2}N{2}{2} @{}} 
      \hline
Train data & Setup & MORPH & POSMORPH \\
      \hline
     PDT & PCRF~\cite{Mueller:2013} & 6.07 & 7.01\\
     & BLSTM~\cite{Ling:2015a}$^{5}$ & & 6.30\\
       \cline{2-4}
       & CNNHighway-BLSTM & 4.48 & 4.87\\
      & LSTM-BLSTM & 3.92 & 4.36\\
      \hline
            + cs.wikidump & CRF + MarLiN~\cite{Mueller:2015} & 5.67 &\\
           \cline{2-4}
           & CNNHighway-BLSTM & 3.79 & 4.19\\
      & LSTM-BLSTM & 3.95 & 4.07\\
      \hline
   \end{tabular}
   \npnoround
   \label{tab:czech}
\end{table*}

%

\clearpage
\newcommand{\SortNoop}[1]{}

\end{document}